\documentclass{bmvc2k}

\usepackage{times}
\usepackage{epsfig}
\usepackage{graphicx}
\usepackage{amsmath}
\usepackage{amssymb}
\usepackage[bb=boondox]{mathalfa}
\usepackage{algorithm}
\usepackage[noend]{algpseudocode}
\usepackage{subcaption}
\usepackage{dblfloatfix}
\usepackage{enumitem}

\usepackage{xfrac}
\usepackage{rotating}
\usepackage{outlines}
\usepackage{multirow, multicol}
\usepackage{booktabs}
\usepackage{adjustbox}
\usepackage{array}

\usepackage{hyperref}

\DeclareCollectionInstance{custom_math}{xfrac}{mathdefault}{math}
{
    slash-right-mkern = -1 mu,
    slash-left-mkern = -1 mu
}
\UseCollection{xfrac}{custom_math}

\title{VIN: Voxel-based Implicit Network for Joint 3D Object Detection and Segmentation for Lidars}

\addauthor{Yuanxin Zhong}{zyxin@umich.edu}{1}
\addauthor{Minghan Zhu}{minghanz@umich.edu}{1}
\addauthor{Huei Peng}{hpeng@umich.edu}{1}

\addinstitution{
 Mechanical Engineering\\
 University of Michigan\\
 Ann Arbor, USA
}

\runninghead{Zhong, Zhu, Peng}{Joint 3D Object Detection and Point Cloud Segment...}

\newcommand{\squeezeup}{\vspace{-3mm}}
\newcolumntype{R}[2]{%
    >{\adjustbox{angle=#1,lap=\width-(#2)}\bgroup}%
    l%
    <{\egroup}%
}
\newcommand*\rot{\multicolumn{1}{R{45}{1em}}}

\definecolor{nuscenes_car}{rgb}{1.0, 0.620, 0.0}
\definecolor{nuscenes_truck}{rgb}{1.0, 0.388, 0.278}
\definecolor{nuscenes_ped}{rgb}{0.0, 0.0, 0.902}
\definecolor{nuscenes_veg}{rgb}{0.0, 0.686, 0.0}
\definecolor{nuscenes_sidewalk}{rgb}{0.294, 0.0, 0.294}
\definecolor{nuscenes_drive}{rgb}{0.0, 0.812, 0.749}
\definecolor{nuscenes_terrain}{rgb}{0.439, 0.706, 0.259}
\definecolor{nuscenes_manmade}{rgb}{0.871, 0.722, 0.529}

\begin{document}

\maketitle

\begin{abstract}
   A unified neural network structure is presented for joint 3D object detection and point cloud segmentation in this paper.  We leverage rich supervision from both detection and segmentation labels rather than using just one of them. In addition, an extension based on single-stage object detectors is proposed based on the implicit function widely used in 3D scene and object understanding. The extension branch takes the final feature map from the object detection module as input, and produces an implicit function that generates semantic distribution for each point for its corresponding voxel center. We demonstrated the performance of our structure on nuScenes-lidarseg, a large-scale outdoor dataset. Our solution achieves competitive results against state-of-the-art methods in both 3D object detection and point cloud segmentation with little additional computation load compared with object detection solutions. The capability of efficient weakly supervised semantic segmentation of the proposed method is also validated by experiments.
\end{abstract}

\section{Introduction}
\label{sec:intro}
3D object detection and scene understanding are two major perception tasks for 3D computer vision. The former task provides position and dimension information for dynamic objects of interest, while the latter task helps to understand the environment, which is usually accomplished by semantic segmentation of the sensor data. These tasks are important for autonomous driving and mapping.  The detected 3D bounding boxes are useful for object behavior prediction, while the semantic information is useful for lane-keeping and static obstacle avoidance. Significant prior works exist using lidar (Light Detection and Ranging) sensors for these tasks thanks to their superior ranging precision and robustness to certain environmental factors such as low/high lighting.

During the past decade, techniques for object detection and segmentation have advanced significantly. Convolutional Neural Networks (CNN) have been widely used, including \cite{duan2019centernet, ren2015faster, tan2020efficientdet} for 2D object detection, \cite{zhao2017pyramid, he2017mask, chen2017deeplab} for 2D semantic segmentation, \cite{zhou2018voxelnet, qi2018frustum, yin2020center} for 3D object detection and \cite{zhong20173d, zhu2020cylindrical} for 3D semantic segmentation.

Few of these works combine the detection and segmentation tasks, several follow the panoptic segmentation scheme proposed by \cite{kirillov2019panoptic}. Examples include \cite{wang2020axial} on 2D images, and \cite{zhu2020cylindrical} on 3D point clouds. The simultaneous multi-task learning strategy is favorable for real-time applications, since it's usually computationally efficient to generate detection and segmentation results simultaneously, as some computations can be re-purposed.

\begin{figure}[!ht]
\centering
\bmvaHangBox{\includegraphics[trim=0 100 0 70,clip,width=0.7\columnwidth]{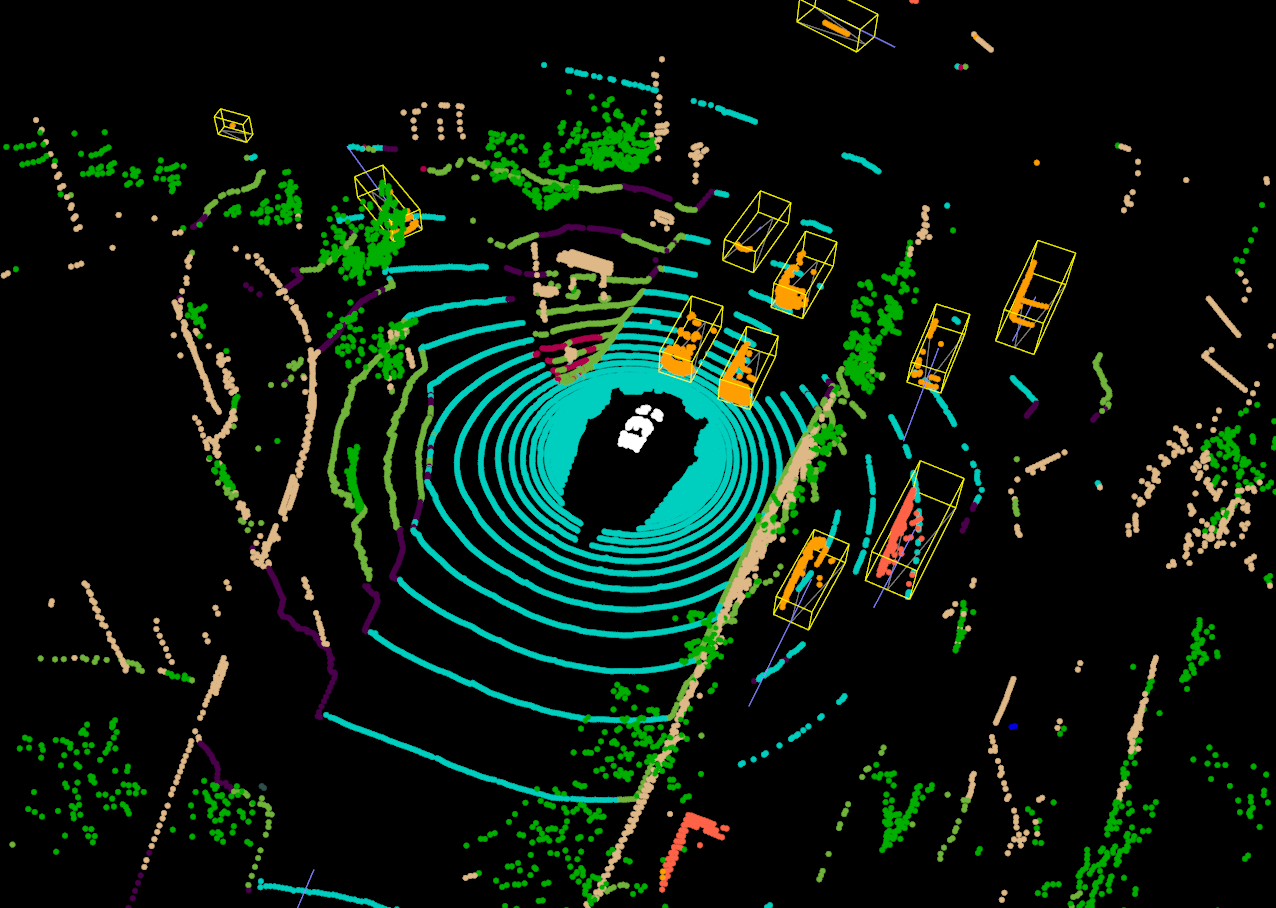}}
   \caption{\small An example of simultaneous object detection and segmentation. \\ \textit{Legend of point clouds: \textcolor{nuscenes_car}{$\blacksquare$ Car}
   \textcolor{nuscenes_truck}{$\blacksquare$ Truck}
   \textcolor{nuscenes_ped}{$\blacksquare$ Pedestrian}
   \textcolor{nuscenes_veg}{$\blacksquare$ Vegetation}
   \textcolor{nuscenes_sidewalk}{$\blacksquare$ Sidewalk}
   \textcolor{nuscenes_drive}{$\blacksquare$ Drivable Surface}
   \textcolor{nuscenes_terrain}{$\blacksquare$ Terrain}
   \textcolor{nuscenes_manmade}{$\blacksquare$ Man-made}
} Best viewed in color and zoomed in.}
\label{fig:thumbnail}
\squeezeup
\end{figure}

It's worth remembering that 3D detection is different from 2D detection in the sense that a 2D bounding box is usually a tight bound of corresponding instance mask, while 3D detection (in outdoor scenes) provides object shape and orientation. Instance segmentation in 2D is also inherently different from 3D instance segmentation in terms of overlap, since there may exist occluded objects in 2D images where pixels of one object may locate inside the bounding box of another object, which is not the case in 3D point clouds, where bounding boxes should not overlap under normal situations. These differences make the design of panoptic segmentation methods different between 2D and 3D data.

In this paper, we propose a method called VIN (\textbf{V}oxel-based \textbf{I}mplicit \textbf{N}etwork), which takes a 3D lidar point cloud as input and reports both 3D object detection and semantic segmentation results as the outputs. To the best of our knowledge, this is one of the first papers that perform lidar-based 3D object detection and point cloud segmentation through a single network. Our main contributions include:
\begin{enumerate}
    \item A semantic branch from a voxel-based 3D object detector which adds little computation overhead for the additional output. The semantic branch can be trained with weak supervision. The performance with just \textbf{0.1\%} semantic labels, after some training, was found to be on par with a model trained with full supervision.
    \item A strategy to fix inconsistency between bounding boxes and point-wise semantic labels, validated by our experiments.
    \item Improved semantic and panoptic segmentation quality compared with state-of-the-art methods based on the results from the nuScenes-lidarseg dataset.
\end{enumerate}

The remainder of the paper is organized as follows: Section \ref{sec:related_work} introduces related work in 3D object detection and segmentation, Section \ref{sec:method} describes the structure of the proposed network, Section \ref{sec:experiment} presents the settings and performance metrics of our experiments. Finally, Section \ref{sec:conclusion} concludes this paper.
\section{Related Work}
\label{sec:related_work}

\subsection{3D Object Detection}

 Due to the wide availability of various datasets including \cite{geiger2013vision, caesar2020nuscenes, dong2019mcity}, 3D object detection has become a hot topic in computer vision. Grouped by the modality of sensors used in the detection, the methods in literature can be categorized as image only \cite{chen2016monocular, mousavian20173d, chabot2017deep, wang2019pseudo}, point cloud only \cite{qi2018frustum, zhou2018voxelnet, yan2018second, yin2020center, shi2020pv} and multi-modal fusion \cite{sindagi2019mvx, meyer2019sensor}. Among the lidar-based detection algorithms, the approaches include feature extraction by points, voxels, projected images, and combinations \cite{shi2020pv, ku2018joint}. The point-based approaches \cite{qi2018frustum} apply Multi-Layer Perceptron (MLP) and gather feature from points in the same group, while the voxel-based methods \cite{zhou2018voxelnet, yin2020center, sindagi2019mvx} first assign the point cloud into voxel grids, and then convolutional neural network (CNN) modules are applied to the voxels. The projection-based methods \cite{meyer2019sensor} project the point cloud into images in a perspective view and then features are extracted using 2D CNN models. After the feature extraction, these methods usually generate 3D bounding boxes in a single-staged or two-staged manner (\cite{liu2016ssd} and \cite{ren2015faster} respectively).

\subsection{Point Cloud Semantic Segmentation}

Point cloud semantic segmentation is an emerging research field taking advantage of datasets including \cite{behley2019semantickitti, caesar2020nuscenes}. Similar to 3D object detection using point clouds, the segmentation algorithms also can use different approaches. PointNet\cite{qi2017pointnet} and its successors \cite{qi2017pointnet++, thomas2019kpconv} directly operate on a point cloud array and report point-wise semantics. Voxel-based frameworks \cite{dai2017scannet, armeni20163d, choy20194d} extract features for voxel grid convolution and reports voxel-wise semantics. Due to the cubic growth of computational cost with the number of voxel grids, sparse convolution \cite{yan2018second, choy20194d} is widely applied for these methods. Voxels are usually sliced in the 3D Cartesian coordinate, while there are also methods \cite{hong2020lidar, zhou2020cylinder3d, zhu2020cylindrical} in the polar or cylindrical coordinate systems. Methods using projected images \cite{wu2018squeezeseg} for semantic segmentation take advantage of well-explored techniques developed for the 2D segmentation.

Different representations for point clouds have respective strengths and drawbacks. Point-wise feature operation preserves the granularity of the original point cloud, but suffers from heavy computation cost. Voxel-wise feature operations are fast, but suffer from lower precision introduced by the rasterization. Projection-based methods can take advantage of handy 2D convolution structures and their efficiency but need to learn to reconstruct the 3D object shape.  In this paper, we enhance the voxel-based framework for its efficiency, by using the implicit representation introduced in Section \ref{sec:method}, to enable precise point-wise predictions.

Recently, many researchers \cite{gojcic2021weakly, mei2019semantic, liu2021one} work to advance the ability of point cloud semantic segmentation with fewer labels, namely weakly or semi-supervised semantic segmentation. Most of them depend on unsupervised algorithms \cite{gojcic2021weakly, liu2021one} to cluster the point cloud, or depends on consistency between data frames \cite{mei2019semantic}. In this paper, we propose a method that can handle weak supervision with the help of object detection results.

\subsection{Joint 3D Detection and Segmentation}

Both 3D object detection and environment segmentation are important tasks for autonomous driving. Relatively few papers combine the two tasks to provide richer information for downstream modules. \cite{meyer2019sensor} achieved joint 3D object detection and point cloud segmentation by combining the features from the lidar point cloud and camera image and extrinsic projection to collect additional point cloud features from the image feature map. \cite{unal2021improving} improved semantic segmentation performance by adding object detection as an auxiliary downstream stage for additional supervision, which shares a similar structure as this paper, but they started from a point-based semantic backbone. In addition, the two approaches cited above only produce results in a field of view limited by the camera. In this paper, we aim to develop detection and segmentation results without this limitation.

There are many other methods \cite{hong2020lidar, zhu2020cylindrical} that produce panoptic segmentation results of a point cloud, which is different from joint 3D detection and segmentation as discussed in Section \ref{sec:intro}. \cite{hong2020lidar} proposed a shifting network module to predict whether a point belongs to a certain instance or not by location regression. \cite{zhu2020cylindrical} proposed a panoptic segmentation framework based on a cylindrical representation of the point cloud augmented by asymmetrical convolutional modules. Compared with these methods, our method achieves panoptic segmentation by generating instance labels with predicted bounding boxes.

\subsection{Implicit Representation}

Many recent computer vision algorithms use 2D or 3D grids which result in loss of data granularity. To preserve precision, a possible solution is to operate on the raw points, and another way is to use implicit representation such as signed distance functions \cite{mescheder2019occupancy, park2019deepsdf}. The key insight behind the implicit representation is to learn a function to represent the input location, instead of learning the collection of per-location predictions given the feature map input. The prediction can be occupancy of the location as used in \cite{mescheder2019occupancy}, or a direct semantic label as what we used in this paper. A major benefit of this approach is that we can predict continuously, i.e. prediction can be obtained for arbitrary query positions, whether it's aligned with the original data or not.
\section{Methodology}
\label{sec:method}

Different from most methods mentioned in Section \ref{sec:related_work}, our goal is to generate 3D bounding boxes and point-wise semantic labels simultaneously given a point cloud input. On top of these results, panoptic segmentation predictions can be generated easily. First, We will describe the backbone applicable to our algorithm in Section \ref{sec:backbone}. Then the details of the proposed semantic branch will be elaborated in \ref{sec:semantic_branch}. The panoptic segmentation will be covered in Section \ref{sec:panoptic}. Figure \ref{fig:framework} illustrates the structure of the proposed method.

\subsection{Backbone Network}
\label{sec:backbone}

\begin{figure*}[!t]
\centering
\includegraphics[width=\textwidth]{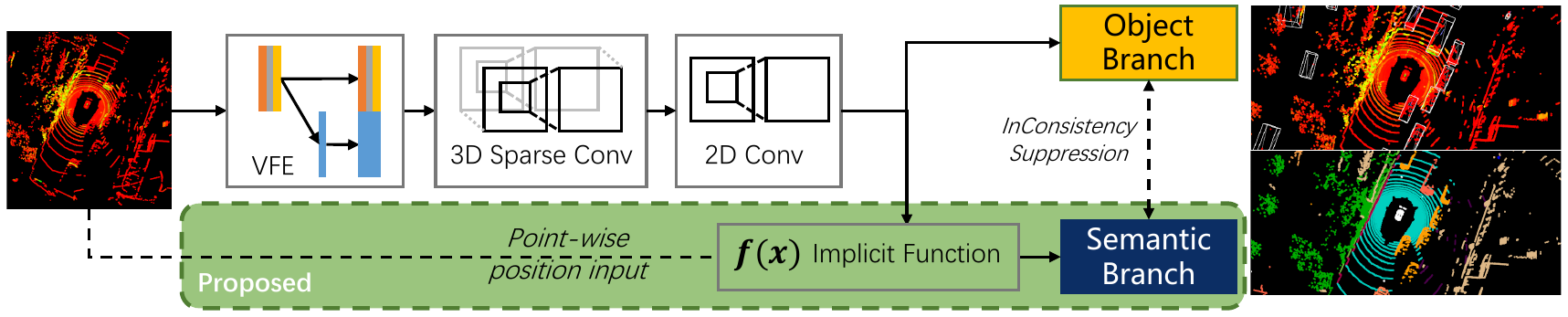}
\caption{The overall structure of the proposed joint 3D object detection and semantic segmentation framework}
\label{fig:framework}
\squeezeup
\end{figure*}

Our method can work with any voxel-based detectors that predict bounding boxes based on a feature map. To show the effectiveness of our semantic branch, we use CenterPoint\cite{yin2020center} as our backbone. CenterPoint follows the VoxelNet\cite{zhou2018voxelnet} structure, consisting of voxel feature extraction (VFE) layers, a 3D convolutional backbone, a bird's eye view 2D convolutional backbone and several detection heads (see the upper branch of Figure \ref{fig:framework}. After object proposals are generated for each location, they are passed through a Non-Maximum Suppression (NMS) or Circular NMS module to collect final predictions.

\subsection{Semantic Rendering Branch}
\label{sec:semantic_branch}

Inspired by \cite{kirillov2020pointrend}, we propose a semantic rendering branch that learns a continuous function to generate semantic distribution prediction for each input position and reduce rasterization error. The semantic branch prevents granularity loss by using an implicit function. This process is achieved by using a lightweight MLP module with inputs coming from both point positions and local voxel features. Different from PointRend which uses both coarse and fine-grained features, we only use the original global feature map from the convolutional backbone since our base framework is a single-stage detector instead of the two-stage MaskRCNN\cite{he2017mask} used in \cite{kirillov2020pointrend}. In our experiment, the MLP of the semantic branch consists of 4 layers with 256, 128, 64, and 32 channels, respectively.

Given a query point $q=(x_q, y_q, z_q)$ from the point cloud input and backbone feature map $\mathbf{M}\in\mathbb{R}^{C\times D\times H\times W}$, we first find the grid position $(i, j, k)\in \mathbb{N}^{D\times H\times W}$ in the feature map where the query point lies, then the semantic distribution $s_q \in \mathbb{R}^S$ with $S$ classes of the point are generated by the MLP module $f: \mathbb{R}^{3+C}\rightarrow\mathbb{R}^{S}$ formulated as \begin{align}s_q = \mathrm{softmax}(f(\left[x_q-cx_i\;y_q-cy_j\;z_q-cz_k\;\mathbf{M}_{\cdot ijk}\right]))\end{align}
where $(cx_i, cy_j, cz_k) \in \mathbb{R}^{D\times H\times W}$ represents the real-world center of the grid in the feature map indexed by $(i,j,k)$. If the query point is outside of the voxel grid, then the voxel closest to the point is selected. This module $f$ will be supervised by the point-wise semantic labels from the dataset.

Since only the position of the query point is fed into the function, unlike the method used in \cite{zhu2020cylindrical}, the semantic branch can predict points that are not in the original point cloud. The decoupling between feature extraction and querying is beneficial when semantics are required to be extrapolated between points, which will be further discussed in Section \ref{sec:utilize_implicit}. Note that in our backbone, the 3D voxel features are flattened and processed by a 2D convolutional network, therefore in our case $D=1$. The proposed branch can be applied directly to a 3D voxel grid if the backbone framework produces a 3D tensor to the detection head.

With the network structure defined above, we define the loss function as \begin{align}L=\alpha_{cls}L_{cls}+\alpha_{reg}L_{reg}+\alpha_{sem}L_{sem}
\end{align}
where we use focal loss with Gaussian kernels for the classification loss $L_{cls}$ on the heatmap, and $L_1$ loss for the regression loss $L_{reg}$ on box parameters, both of which are inherited from the backbone method we adopted from \cite{yin2020center}. For semantic supervision, we use a combination of Lovász oss \cite{berman2018lovasz} and weighted cross-entropy loss for the semantic loss $L_{sem}$.In our experiment, the parameters for the weight loss are $\alpha_{cls}=\alpha_{sem}=1,\alpha_{reg}=0.25$.


\subsection{Panoptic Post-processing}
\label{sec:panoptic}

After bounding boxes and point-wise semantic labels are generated, panoptic segmentation results can be generated by assigning instance ids of the boxes to the points inside them.

Aside from panoptic label generation, object bounding boxes and semantic labels can be used to mutually recover the error in predictions. Inspired by the strategy introduced in \cite{zhong20173d}, we developed a novel procedure named \textbf{InConsistency Suppression} (ICS) to first fix inconsistent labels of the bounding boxes using a estimated label from point-wise semantic outputs and then inconsistent points will be fixed. The algorithm is described in the pseudocode (Algorithm \ref{alg:ics}) below. The inputs for the procedure are the collection of predicted bounding boxes $B$ and semantic point clouds $P$. Each box $b\in B$ has an object classification $K(b)$ with confidence score $s(b)$, and each point $p \in P$ has semantic classification $K(p)$ with confidence score $s(p)$. $b$ also denotes the area inside the box. In the procedure description, $K_{\text{th}}$ denotes the thing categories.

\begin{algorithm}[!h]
\caption{InConsistency Suppression}
\label{alg:ics}
\begin{algorithmic}[1]

\Procedure{ICS}{$B,P$}       \Comment{Fix inconsistent labels in-place}
    \Statex $B$: labeled bounding boxes, $P$: labeled point cloud
    \Statex $c_\alpha$, $c_\gamma$: tunable parameters with default value 1
    \Statex $m_p$: score margin for overriding point label
    \Statex
    \State Sort $B$ descendingly by score
    \For{$i=1\ldots|B|$} \Comment{loop for fixing box label}
        \State $P_i\leftarrow\{p\in P \cap b_i| K(p)\neq K(b_j)\;\forall j<i,p\in b_j\}$  \Comment{select inconsistent points}
        \For{$k\in K_{\text{th}}$} \Comment{for each semantic class $k$}
            \State $P_i^k\leftarrow \{p\in P_i|K(p)=k\}$ \Comment{select points with label $k$}
            \State $\alpha_k\leftarrow \sfrac{|P_i^k|}{|P_i|}$ \Comment{count criterion}
            \State $\beta_k\leftarrow \sfrac{\sum_{p\in P_i^k} s(p)}{|P_i^k|}$ \Comment{score criterion}
            \State $\gamma_k\leftarrow 1+s^{c_\gamma} (b_i)\cdot\mathbb{1}_{k=K(b_i)}$ \Comment{correctness criterion}
        \EndFor
        \State $k^*\leftarrow\arg\max_{k\in K_{\text{th}}} \alpha_k\beta_k\gamma_k$ \Comment{select the best class using the three criteria}
        \If{$k^*\neq K(b_i)$}  \Comment{if the best class is not the predicted one}
            \If{$\exists\,j>i\;\text{s.t.}\;K(b_j)=k^*$ }
                \State Swap $K(b_i)$ and $K(b_j)$ \Comment{swap label with a box with lower score}
            \Else
                \State $K(b_i)\leftarrow k^*$ \Comment{override the box label by the best class}
            \EndIf
        \EndIf
    \EndFor
    \For{$i=|B|\ldots 1$} \Comment{loop for fixing point labels}
        \For{$p \in \{p\in P \cap b_i| s(p) < s(b_i) - m_p\}$} \Comment{for each point with low score}
            \If{$\exists\,j<i\;\text{s.t.}\;p \in b_i\cap b_j$}
                \State \textbf{Continue} \Comment{ignore boxes with large overlap}
            \EndIf
            \State $K(p)\leftarrow K(b_i)$  \Comment{override the point label by box label}
        \EndFor
    \EndFor
\EndProcedure

\end{algorithmic}
\end{algorithm}
\squeezeup\squeezeup

\section{Experiment Results}
\label{sec:experiment}

Experiments are conducted on the nuScenes\cite{caesar2020nuscenes} dataset with nuScenes-lidarseg extension. This dataset is selected because it provides labels for both 3D object detection and point cloud semantic segmentation.  The nuScenes dataset contains various labels for different tasks, including 28k synchronized frames with multiple cameras, one lidar and multiple radars for training, and 6k samples each for validation and testing. The nuScenes-lidarseg extension provides point-wise semantic labels of 15 categories in total. It's a challenging dataset with adverse environment scenarios including dark nights and rainy days.

Our backbone method is CenterPoint\cite{yin2020center} with a voxel size of 0.1m and without CBGS\cite{zhu2019class} or other test-time augmentation due to our hardware limitation. Our model is trained with 2 NVidia V100 16G GPUs, using AdamW optimizer with cyclic learning rate scheduling starting from $1e^{-4}$ and weight decay of 0.01. Please refer to \cite{yin2020center} and our code for details.

\begin{figure*}[!t]
\centering
\includegraphics[width=\textwidth]{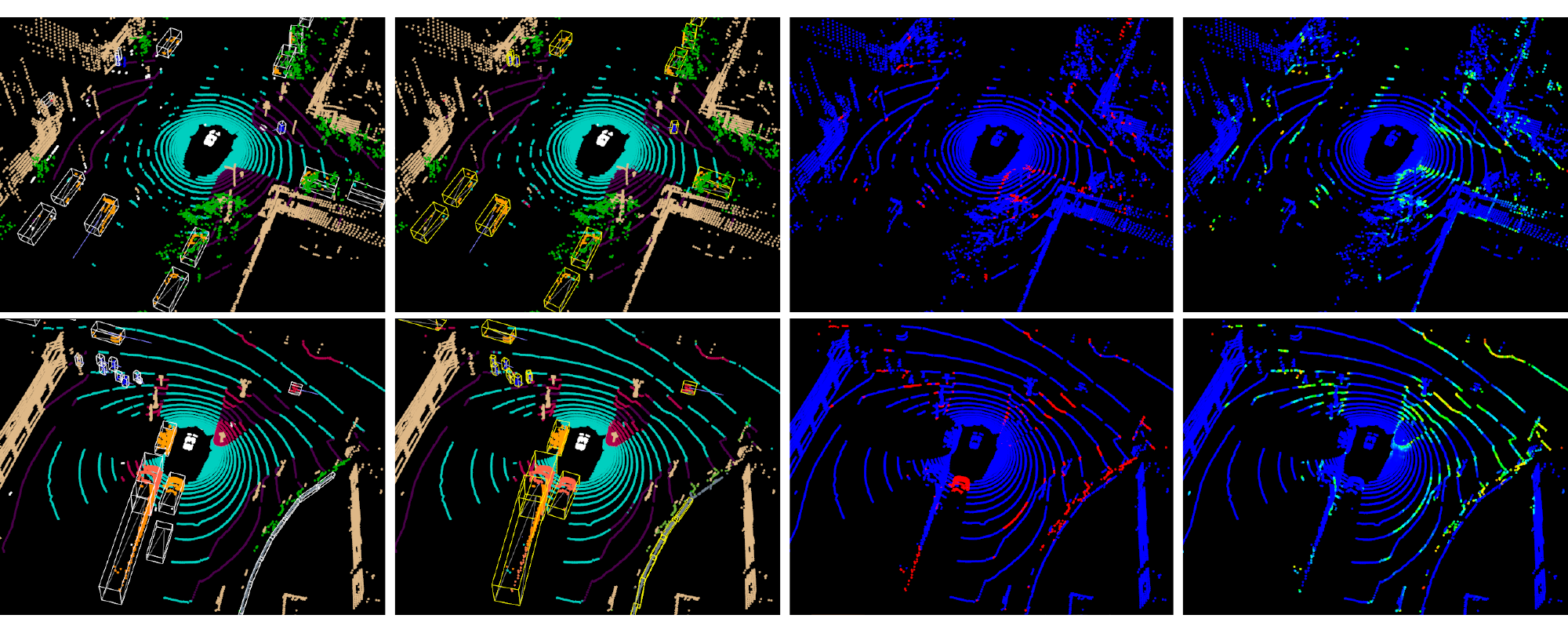}
\caption{\small Qualitative results of VIN. First column: ground-truth semantic labels and bounding boxes; Second column: estimated semantic labels and bounding boxes; Third column: error points of semantic segmentation (labeled in red); Forth column: confidence score of the semantic segmentation (colors vary from light green to dark blue for scores from low to high). Best viewed in color and zoomed in.}
\label{fig:examples}
\end{figure*}

Selected qualitative results are shown in Figure \ref{fig:examples}. Our method is able to generate precise point-wise semantic labels while preserving the ability to predict accurate 3D bounding boxes. However, our method still suffers from the problem of ambiguity due to the sparsity of the lidar point cloud, common for all lidar-based detection or segmentation algorithms.

\subsection{Quantitative Results}

\textbf{Semantic Segmentation Performance}
For segmentation performance, our method is compared against  state-of-the-art methods on the nuScenes lidar segmentation track. The main metrics used for comparison are the intersection-over-union (IoU) for each category. Mean IoU (mIoU) and frequency-weighted IoU (fwIoU) numbers are also provided by the nuScenes benchmark. The results are presented in Table \ref{tab:semantic_results}, compared with other methods, our approach performs significantly better in most "things" categories except for the traffic cones. Our backbone method does not preserve details in each voxel, thus it may not work well with small objects. On the other hand, existing methods outperform in many "stuff" categories, because our network needs to focus on objects to learn object detection well. Overall, our method outperforms the state-of-the-art methods by about 9.4 percentage points in mIoU.

\textbf{3D Object Detection Performance}:
For the lidar-only object detection performance, we select a few other state-of-the-art methods on the nuScenes detection track. The overall performance is measured by mean average precision (mAP) and the NDS score proposed by nuScenes to capture not only precision performance, but also errors of other target properties. From the results shown in Table \ref{tab:detection_results}, we found that although our method suffers performance loss when adding a semantic branch, we still achieve good performance in most categories. This indicates that our method can produce semantic segmentation labels without much loss of detection performance. The most prominent gaps come from the "trailer", "bicycle" and "construction vehicle" categories, all of which are rare in the dataset. It's hard for the network to maintain the same performance when it learns to solve an additional segmentation task. Label balancing techniques can be applied in the future to solve the problem.

\begin{table*}[t]
    \centering
    \captionsetup{justification=centering}
    \caption{\small Comparison of semantic segmentation performance using the nuScenes validation dataset. For all metrics the higher the better, the best one is shown in boldface.}
    \resizebox{\textwidth}{!}{
    \footnotesize
    \begin{tabular}{l|c|cc|ccccccccccccccc}
        \toprule
        Method & Source & mIoU & fwIoU & \rot{Car} & \rot{Truck} & \rot{Bus} & \rot{Trailer} & \rot{\shortstack[l]{Construction\\Vehicle}} & \rot{Pedestrian} & \rot{Motorcycle} & \rot{Bicycle} & \rot{Traffic Cone} & \rot{Barrier} & \rot{Drivable Area} & \rot{Other flat} & \rot{Terrain} & \rot{Man-made} & \rot{Vegetation} \\
        \midrule
        SalsaNext \cite{cortinhal2020salsanext} & \cite{cheng20212} & 58.8 & 82.8 & 81.0 & 65.7 & 77.1 & 38.3 & 18.4 & 52.8 & 47.5 & 4.7 & 43.5 & 56.6 & 94.2 & 60.0 & \textbf{70.3} & 81.2 & 80.5 \\
        MinkNet42 \cite{choy20194d} & \cite{cheng20212} & 60.8 & 82.7 & 77.1 & 62.2 & 77.4 & 42.5 & 23.0 & 55.6 & 55.1 & 8.3 & 50.0 & 63.1 & 94.0 & 67.2 & 68.6 & \textbf{83.7} & 80.8\\
        (AF)\textsuperscript{2}-S3Net \cite{cheng20212} & \cite{cheng20212} & 62.2 & 83.0 & 80.0 & 67.4 & 82.3 & 42.2 & 20.1 & 59.0 & 62.0 & 12.6 & 49.0 & 60.3 & 94.2 & 68.0 & 68.6 & 82.9 & \textbf{82.4} \\
        
        \midrule

        VIN (ours) & - & 73.7 & 84.3 & \textbf{87.0} & 82.6 & \text{91.7} & 63.2 & 49.9 & 79.7 & 82.2 & \textbf{46.2} & \textbf{59.4} & 74.7 & 94.5 & 67.1 & 68.5 & \textbf{83.7} & 81.2 \\
        VIN + ICS (ours) & - & \textbf{73.8} & 84.4 & \textbf{87.0} & \textbf{82.8} & 91.7 & \textbf{63.7} & \textbf{50.4} & 79.8 & \textbf{82.5} & 46.1 & \textbf{59.4} & 74.7 & 94.5 & 67.1 & 68.5 & \textbf{83.7} & 81.2 \\
        VIN (seg only) & - & 72.0 & \textbf{86.3} & 86.6 & 82.0 & \textbf{91.8} & 61.8 & 45.9 & \textbf{80.3} & 68.6 & 30.7 & 45.2 & \textbf{75.4} & \textbf{95.8} & \textbf{73.8} & \textbf{72.6} & \textbf{85.3} & \textbf{83.3} \\

        \bottomrule
    \end{tabular}
    }
    \label{tab:semantic_results}
    \squeezeup
\end{table*}


\begin{table*}[t]
    \centering
    \captionsetup{justification=centering}
    \caption{\small Comparison of (lidar-only) 3D object detection performance on the nuScenes validation dataset. All metrics are the higher the better, the best one is underlined. \\ Abbreviations: \textit{CV - construction vehicle, TC - traffic cone, Ped - Pedestrian, Motor - Motorcycle, R - Reproduced}\\}
    \resizebox{\columnwidth}{!}{
    \footnotesize
    \begin{tabular}{l|c|cc|cccccccccc}
        \toprule
        Method & Source & mAP & NDS & Car & Truck & Bus & Trailer & CV & Ped & Motor & Bicycle & TC & Barrier \\
        \midrule
        PointPillars \cite{lang2019pointpillars} & \cite{zhang2020multi} & 28.2 & 46.8 & 75.5 & 31.6 & 44.9 & 23.7 & 4.0 & 49.6 & 14.6 & 0.4 & 8.0 & 30.0 \\
        3DSSD \cite{yang20203dssd} & \cite{zhang2020multi} & 42.6 & 56.4 & 81.2 & 47.2 & 61.4 & 30.5 & 12.6 & 70.2 & 36.0 & 8.6 & 31.1 & 47.9 \\
        CenterPoint \cite{yin2020center} & \cite{zhang2020multi} & 56.6 & 65.0 & 84.6 & 54.7 & 66.0 & 32.3 & 15.1 & 84.5 & 56.9 & 38.6 & 67.4 & 66.1 \\
        
        \midrule
        CenterPoint & R & \underline{50.7} & \underline{60.2} & \underline{82.6} & \underline{50.3} & \underline{63.8} & \underline{30.8} & \underline{12.0} & \underline{79.9} & 43.9 & \underline{22.3} & \underline{60.8} & \underline{60.4} \\
        VIN (Ours) & - & 45.2 & 57.0 & 82.1 & 48.3 & 61.3 & 18.7 & 3.5 & 73.4 & \underline{50.4} & 2.2 & 56.1 & 55.9 \\
        VIN + ICS (Ours) & - & 46.3 & 57.6 & 82.1 & 48.1 & 61.1 & 22.7 & 7.2 & 74.3 & 50.4 & 4.1 & 56.6 & 56.4 \\

        \bottomrule
    \end{tabular}
    }
    \label{tab:detection_results}
    \squeezeup
\end{table*}

\begin{table*}[t]
    \centering
    \caption{\small Comparison of panoptic segmentation performance on nuScenes validation set. All metrics are the higher the better, the best ones are highlighted in boldface.}
    \resizebox{\columnwidth}{!}{
    \footnotesize
    \begin{tabular}{l|c|cccc|ccc|ccc|c}
        \toprule
        Method & Source & PQ & PQ\textsuperscript{$\dagger$} & RQ & SQ & PQ\textsuperscript{th} & RQ\textsuperscript{th} & SQ\textsuperscript{th} & PQ\textsuperscript{st} & RQ\textsuperscript{st} & SQ\textsuperscript{st} & mIoU  \\
        \midrule
        Cylinder3D\cite{zhou2020cylinder3d} + SECOND\cite{yan2018second} & \cite{hong2020lidar} & 40.1 & 48.4 & 47.3 & \textbf{84.2} & 29.0 & 33.6 & \textbf{84.4} & 58.5 & 70.1 & 83.7 & 58.5 \\
        Cylinder3D\cite{zhou2020cylinder3d} + PointPillars\cite{lang2019pointpillars} & \cite{hong2020lidar} & 36.0 & 44.5 & 43.0 & 83.3 & 23.3 & 27.0 & 83.7 & 57.2 & 69.6 & 82.7 & 52.3 \\
        DS-Net\cite{hong2020lidar} & \cite{hong2020lidar} & 42.5 & 51.0 & 50.3 & 83.6 & 32.5 & 38.3 & 83.1 & 59.2 & 70.3 & \textbf{84.4} & 70.7 \\
        
        \midrule

        VIN (ours) & - & \textbf{51.7} & \textbf{57.4} & 61.8 & 82.6 & \textbf{45.7} & 53.7 & 83.6 & \textbf{61.8} & \textbf{75.4} & 80.9 & 73.7 \\
        VIN + ICS (ours) & - & \textbf{51.7} & \textbf{57.5} & \textbf{61.9} & 82.6 & \textbf{45.7} & \textbf{53.8} & 83.6 & \textbf{61.8} & \textbf{75.4} & 80.9 & \textbf{73.8} \\

        \bottomrule
    \end{tabular}
    }
    \label{tab:panoptic_results}
    \squeezeup
\end{table*}

\begin{table*}[t]
    \centering
    \captionsetup{justification=centering}
    \caption{\small Comparison of detection and segmentation performance on nuScenes validation set with different supervision levels. The label percentages denote the ratio of points used in training the semantic branch. All metrics are higher the better.}
    \resizebox{0.9\columnwidth}{!}{
    \footnotesize
    \begin{tabular}{l|cc|cccc|cccc}
        \toprule
        Method & mAP & NDS & mIoU & fwIoU & mIoU\textsuperscript{th} & mIoU\textsuperscript{st} & PQ & PQ\textsuperscript{$\dagger$} & PQ\textsuperscript{th} & PQ\textsuperscript{st} \\
        \midrule
        
        Ours (full supervision) & 45.2 & 57.0 & 73.7 & 84.3 & 71.3 & 78.3 & 51.7 & 57.4 & 45.7 & 61.8 \\
        Ours (10\% label)       & 45.7 & 57.4 & 74.4 & 84.8 & 72.3 & 78.0 & 52.7 & 58.2 & 46.3 & 63.4 \\
        Ours (1\% label)        & 44.7 & 56.6 & 72.6 & 84.6 & 69.5 & 77.9 & 50.4 & 55.9 & 42.8 & 63.1 \\
        Ours (0.1\% label)      & 45.1 & 57.0 & 72.3 & 84.5 & 69.1 & 77.5 & 50.6 & 56.0 & 43.2 & 62.9 \\
        Ours (0.02\% label)      & 44.5 & 56.6 & 70.6 & 83.8 & 67.1 & 76.5 & 49.0 & 54.5 & 41.3 & 61.8 \\

        \bottomrule
    \end{tabular}
    }
    \label{tab:weak_supervision}
    \squeezeup
\end{table*}

\textbf{Panoptic Segmentation Performance}
Panoptic segmentation is a side product of our method, however, it can be used to evaluate the combined performance of detection and segmentation. The metrics proposed in \cite{kirillov2019panoptic} is leveraged in our experiment, which includes Panoptic Quality (PQ), Segmentation Quality (SQ) and Recognition Quality (RQ). SQ and RQ are analog to point-averaged and instance-averaged IoU. We also adopt the replaced Panoptic Quality (PQ\textsuperscript{$\dagger$}) metric proposed by \cite{hong2020lidar}. The performance of our algorithms is compared with three baseline methods in Table \ref{tab:panoptic_results}. Our algorithms outperform these baseline methods in most metrics except SQ. A possible fix is adding penalty terms for "stuff" points inside object boxes and "thing" points outside object boxes in future work, thus explicitly let the network distinguish between the environment and objects. 

\subsection{Ablation Study}
\textbf{Inconsistency Suppression} The proposed ICS procedure is used to eliminate inconsistency between bounding boxes and semantic labels, which is beneficial for downstream perception modules. This is demonstrated in Table \ref{tab:semantic_results}, \ref{tab:detection_results} and \ref{tab:panoptic_results}. For fixing semantic and panoptic segmentation labels, we set $m_p=0.1$ in Algorithm \ref{alg:ics} and achieve slightly better results. On the other hand, there is a large effect by ICS on detection performance, especially in the `barrier', `construction vehicle' and `bicycle' categories, which indicate that ICS is effective. The improvement comes from using point predictions to help determine the object classification.

\textbf{Weakly supervised segmentation}  Weak supervision can help to reduce label efforts when building datasets. Thanks to the nature of the implicit function representation, it doesn't require the full point cloud for supervision. Our method can handle weakly supervised segmentation tasks by feeding fewer labels during the training process. A key difference between our proposed method and single-task segmentation methods is that our method only get auxiliary information from the bounding box labels. The efficacy of our method is illustrated in Table \ref{tab:weak_supervision}. It can be seen that our method achieves robust segmentation performance even with just 0.1\% semantic labels. However, the performance is not increased monotonically with the amount of available labels, which is unexpected and requires further experiments to explain.

\subsection{Inference Efficiency}

A major motivation to combine detection and segmentation in a single network is to reduce inference time and achieve better real-time efficiency when deployed. We compare our method with baseline CenterPoint in terms of inference time. Measured on a single NVidia 2080Ti graphics card, the original CenterPoint network with 0.1m voxel size achieves 6.0 FPS, while our method with the same voxel size and backbone configuration achieves 5.9 FPS but with the additional semantic segmentation results. This is much better than having a separate segmentation implementation.  For example, a state-of-the-art method \cite{cheng20212} alone takes 0.27 seconds (reported on nuScenes leaderboard, on an NVidia Tesla V100). When deploying our method on a vehicle, further optimization, including network distillation, refactoring using a highly efficient inference framework (e.g. TensorRT) can be implemented.

\subsection{Utilizing the Implicit Function}
\label{sec:utilize_implicit}

Thanks to the fact that our semantic branch is merely based on the feature map produced by the convolutional backbone, semantics queries can be conducted at arbitrary positions in the space, as mentioned in Section \ref{sec:semantic_branch}.

Here we demonstrate two use cases of the query ability, semantic prediction on down-sampled point cloud and dense semantic map generation. The experiment results for the former case are reported in table \ref{tab:downsample_performance}. The down-sampling strategy is useful for reducing on-board latency or helping generate off-board labels for autonomous vehicles. In this experiment, only partial points (down-sampled from the original point cloud) are fed into the trained detection backbone and the semantics of the remainder of the original point cloud is predicted by either the semantic branch or nearest neighbor querying. It can be concluded from the experiment that our model can better capture the semantics in region with no lidar measurements when the original point cloud is down-sampled to reduce the inference time.

\begin{figure*}[!t]
\centering
\includegraphics[width=\textwidth]{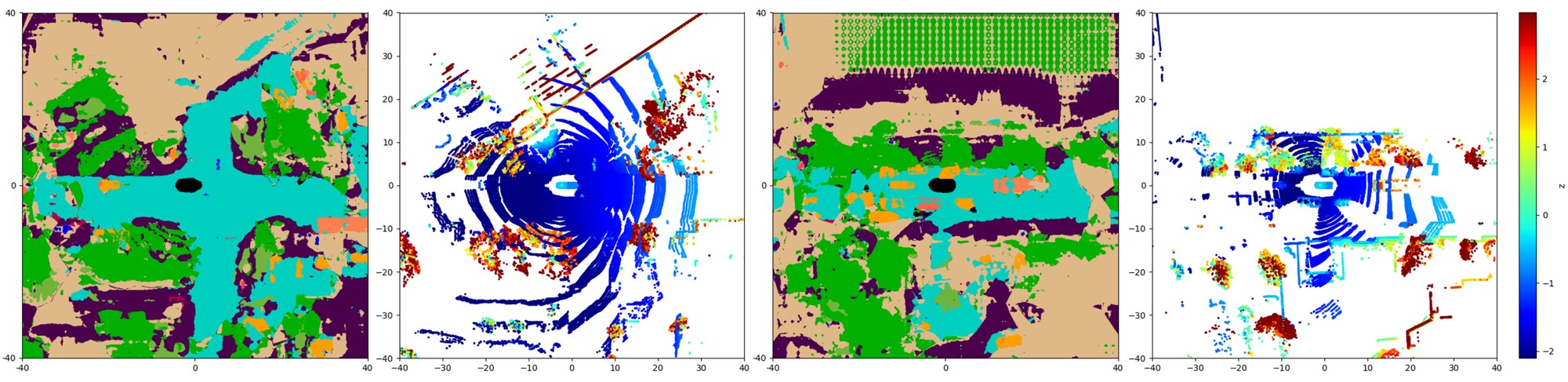}
\caption{\small Two examples of generated dense semantic map on the Nuscenes dataset. The input point cloud is colored by the height. Please refer to Figure \ref{fig:examples} for the color legend.}
\label{fig:dense_sem_map}
\squeezeup
\end{figure*}

On the other hand, it's also possible to generate a dense semantic map by querying semantics at grid points with the semantic branch. Figure \ref{fig:dense_sem_map} shows the semantic prediction of the scenes at certain height (the bilinearly interpolated height of the point cloud is used to create the figure). This dense map can be used as a bird's eye view semantic map, which standalone models have been proposed in the literature (e.g. \cite{reiher2020sim2real}) to estimate. It's useful for finding the semantics boundary for different areas and better interpreting the performance of the deep learning model.

\begin{table*}[t]
    \centering
    \captionsetup{justification=centering}
    \caption{\small Comparison of segmentation performance with partial point cloud input on Nuscenes. The best ones are highlighted in boldface.}
    \resizebox{0.7\columnwidth}{!}{
    \footnotesize
    \begin{tabular}{l|ccc|ccc}
        \toprule
        \multirow{2}{*}{Method} & \multicolumn{3}{c|}{mIoU (random sample)} & \multicolumn{3}{c}{mIoU (beam sample)} \\
        & 100\% & 75\% & 50\% & 32 beams & 24 beams & 16 beams \\
        \midrule
        Ours & 73.7 & \textbf{73.3} & \textbf{72.5} & 73.7 & \textbf{72.5} & \textbf{70.2} \\
        Nearest Neighbor & 73.7 & 73.1 & 72.2 & 73.7 & 64.6 & 55.9 \\
        \midrule
        \textit{(Inference time)} & \textit{169ms} & \textit{133ms} & \textit{104ms} & \textit{169ms} & \textit{125ms} & \textit{96ms} \\
        \bottomrule
    \end{tabular}
    }
    \label{tab:downsample_performance}
    \squeezeup
\end{table*}

\section{Conclusion}
\label{sec:conclusion}

In this paper, a novel framework for joint 3D object detection and semantic segmentation using lidar point clouds is proposed, which is more efficient than two separate implementations. A semantic branch learning implicit representation of spatial semantic properties is proposed. This modification can be applied to any voxel-based 3D object detectors. Experiments using the nuScenes dataset confirm the efficacy and efficiency of our algorithm by comparing it with state-of-the-art methods on semantic and panoptic segmentation tasks.

Currently, our method still struggles to detect and segment partially occluded objects and small objects with few lidar reflections, which are the strengths of camera images, but lidar-based algorithms work better in dazzling and dark scenarios. Future work can incorporate image data to improve the accuracy of both detection and segmentation of the framework.

\section*{Acknowledgements}
This research was funded by Mcity, University of Michgan, under the project "Object Detection and Tracking using Infrastructure and Vehicle-based Sensors."

\bibliography{egbib}
\end{document}